\documentclass[conference]{IEEEtran}

\usepackage[utf8]{inputenc}

\usepackage{tikz}
\usetikzlibrary{positioning}
\usepackage{graphicx}
\IEEEoverridecommandlockouts

\usepackage{color, soul}
\newcounter{LINumberOfComments}
\stepcounter{LINumberOfComments}

\newcounter{SANumberOfComments}
\stepcounter{SANumberOfComments}

\newcommand{\etal}{\textit{et~al.}}

\newcommand{\one}{({\em i})}
\newcommand{\two}{({\em ii})}
\newcommand{\three}{({\em iii})}

\begin{document}

\title{Privacy-Preserving Personal Model Training\\
  \thanks{Supported by EPSRC grants EP/N028260/2 (Databox) and EP/N028422/1 (Contrive). Work completed while Hamed Haddadi and Sandra Servia-Rodríguez were at Queen Mary University of London.} }

\author{\IEEEauthorblockN{1\textsuperscript{st} Given Name Surname}
\IEEEauthorblockA{\textit{dept. name of organization (of Aff.)} \\
\textit{name of organization (of Aff.)}\\
City, Country \\
email address}
\and
\IEEEauthorblockN{2\textsuperscript{nd} Given Name Surname}
\IEEEauthorblockA{\textit{dept. name of organization (of Aff.)} \\
\textit{name of organization (of Aff.)}\\
City, Country \\
email address}
\and
\IEEEauthorblockN{3\textsuperscript{rd} Given Name Surname}
\IEEEauthorblockA{\textit{dept. name of organization (of Aff.)} \\
\textit{name of organization (of Aff.)}\\
City, Country \\
email address}
\and
\IEEEauthorblockN{4\textsuperscript{th} Given Name Surname}
\IEEEauthorblockA{\textit{dept. name of organization (of Aff.)} \\
\textit{name of organization (of Aff.)}\\
City, Country \\
email address}
\and
\IEEEauthorblockN{5\textsuperscript{th} Given Name Surname}
\IEEEauthorblockA{\textit{dept. name of organization (of Aff.)} \\
\textit{name of organization (of Aff.)}\\
City, Country \\
email address}
}

\author{\IEEEauthorblockN{Sandra Servia-Rodríguez\IEEEauthorrefmark{1},
Liang Wang\IEEEauthorrefmark{1},
Jianxin R. Zhao\IEEEauthorrefmark{1},
Richard Mortier\IEEEauthorrefmark{1} and
Hamed Haddadi\IEEEauthorrefmark{2}}
\IEEEauthorblockA{\IEEEauthorrefmark{1}Department of Computer Science and Technology, University of Cambridge}
\IEEEauthorblockA{\IEEEauthorrefmark{2}Dyson School of Design Engineering, Imperial College London}
}

\maketitle

  \begin{abstract}
      Many current Internet services rely on inferences from models trained on user data. Commonly, both the training and inference tasks are carried out using cloud resources fed by personal data collected at scale from users. Holding and using such large  collections of personal data in the cloud creates privacy risks to the data subjects, but is currently required   for users to benefit from such services. We explore how to provide for model training and inference in a system where computation is pushed to the data in preference to moving data to the cloud, obviating many current privacy risks. Specifically, we take an initial model learnt from a small set of users and retrain it locally using data from a single user. We evaluate on two tasks: one supervised learning task, using a neural network to recognise users' current activity from accelerometer traces; and one unsupervised learning task, identifying topics in a large set of documents. In both cases the accuracy is improved. We also analyse the robustness of our approach against adversarial attacks, as well as its feasibility by presenting a performance evaluation on a representative resource-constrained device (a Raspberry Pi).
 \end{abstract}

\begin{IEEEkeywords}
  Privacy, Machine Learning, Algorithms
\end{IEEEkeywords}

\section{Introduction}
\label{sec:introduction}

Large-scale data collection from individuals is at the heart of many current Internet business models. Access to these data allow companies to train models from which to infer user behaviour and preferences, typically leveraging the generous computation resources available in the public cloud. Unfortunately, this data collection is increasingly pervasive and invasive, notwithstanding regulatory frameworks such as the EU's General Data Protection Regulation (GDPR) which attempt to restrain it. The result is that user privacy is compromised, and this is becoming an increasing concern due to reporting of the ongoing stream of security breaches that result in malicious parties accessing such personal data.

Such data collection causes privacy to be compromised even without security being breached though. For example, consider wearable devices that report data they collect from in-built sensors, e.g.,~accelerometer traces and heart rate data, to the device manufacturer. The device might anonymise such data for the manufacturer to use in improving their models for recognising the user's current activity, an entirely legitimate and non-invasive practice. However, the manufacturer might fail to effectively anonymise these data and instead use them for other purposes such as determining mood, or even to sell to third-parties without the users' knowledge. It is not only data from wearables that creates such risks: web queries, article reads and searches, visits to shopping sites and browsing online catalogues are also indexed, analysed, and traded by thousands of tracking services in order to build preference models~\cite{falahrastegar2016tracking}.

So far, users' personal data were mainly sensed through their computers, their smartphones or wearable devices such as smartwatches and smart wristbands. But nowadays smart technology is entering into our homes. We are heading towards an ecosystem where sooner or later, every device in our home will talk to an \emph{Amazon Echo}~\cite{echo}, \emph{Google Home}~\cite{googleHome}, or \emph{Apple HomeKit}~\cite{homeKit}. Apart from controlling smart home appliances such as light bulbs and thermostats with our voice, these smart controllers for the entire home will be required to perform more complex tasks such as detecting how many people are in the house and who they are, recognising the activity they are performing or even telling us what to wear. In this new scenario, users are becoming progressively more aware of the privacy risks of sharing their voice, video or any other data sensed in their homes with the service providers, at the same time that these applications are demanding more accurate and personalised solutions. Sending personal data to the public cloud to perform these tasks seems no longer to be an acceptable solution, but solutions should take advantage of the resource capabilities of personal devices and bring the processing locally, where data resides. 

Approaches such as homomorphic encryption allow user data to be encrypted, protecting against unintended release of such data, while still being amenable to data processing. This affords users better privacy -- their data cannot be used arbitrarily -- while allowing data processors to collect and use such data in cloud computing environments. However, current practical techniques limit the forms of computation that can be supported. We are interested in an alternative approach where we reduce or remove the flow of user data to the cloud completely, instead moving computation to where the data already resides under the user's control~\cite{Chaudhry:2015:PDT:2882850.2882858, databox}. This can mitigate risks of breach and misuse of data by simply avoiding it being collected at scale in the first place: attack incentives are reduced as the attacker must gain access to millions of devices to capture data for millions of users, rather than accessing a single cloud service. However, it presents challenges for the sorts of model learning processes required: how can such models be learnt without access to the users' personal data?



In this paper we address these challenges using the \emph{Edge Computing} paradigm. Specifically, our contributions include: \one~we develop our \emph{personal training} method for implementing machine learning in an environment where personal data largely remains on constrained devices under the control of the data subject~(\S\ref{sec:methodology}); \two~we apply this method to two well-known learning tasks, one supervised (activity recognition from accelerometer traces,~\S\ref{sec:AR}), and one unsupervised (modelling topics in text documents,~\S\ref{sec:topicModelling}) and report the results; and \three~we explore the robustness of our method against adversarial attacks, as well as the feasibility of implementing such techniques on a representative resource-constrained personal device: a Raspberry Pi 3 Model B~\cite{raspberrypi}~(\S\ref{sec:discussion}).

The essence of our approach is a two-step process: \one~we first train a \emph{shared model} using a small set of voluntarily shared user data and distribute this model to all users; and \two~we then retrain this model locally using personal data held by each user, drawing inferences from the resulting \emph{personal model}. We evaluate this approach using \one~a neural network to recognise users' activity on the \emph{WISDM} dataset~\cite{kwapisz2011activity} and \two~the Latent Dirichlet Algorithm (LDA)~\cite{Blei:2003:LDA:944919.944937} to identify topics in the Wikipedia and NIPS datasets~\cite{nips,wikipedia}. In both cases we show that the model resulting from local retraining of an initial model learnt from a small set of users performs with higher accuracy than either the initial model alone or a model trained using only data from the specific user of interest.

We also demonstrate the feasibility of training and testing a small classifier in a resource-constraint, light-weight personal device: a Raspberry Pi 3 Model B~\cite{raspberrypi}. We find that such a device is certainly capable of supporting these algorithms, with negligible time to obtain inferences (on the order of milliseconds), and reasonable training times as well (on the order of tens of seconds).

\section{Methodology}
\label{sec:methodology}

The current approach, which we wish to avoid, of sending all users' personal data to the cloud for processing, is one extreme of a spectrum whose other extreme would be to train a model for a specific user using only that user's data. For some applications, e.g.,~activity recognition, it has been shown that a model trained solely using data from the individual concerned provides more accurate predictions for that individual than a model trained using data from other individuals~\cite{weiss2012impact}. At the same time, this solution offers more privacy to the user as all computation, for both training and inference, can be done locally on the device~\cite{Chaudhry:2015:PDT:2882850.2882858}. However, this approach leads to substantial interactional overheads as training the model will likely require each user to label a significant amount of data by hand before they can obtain accurate inferences.

We propose and evaluate an alternative, hybrid approach that splits computation between the cloud and the users' personal devices. We start by first training a model in the cloud using data from a small (relative to the population) set of users. We then distribute this \emph{shared model} to users' personal devices, where it can be used locally to generate inferences. In addition, it can be retrained using locally-stored personal data to become a \emph{personal model}, specialised for the user in question.

\begin{figure}[t!]
  \centering
  \includegraphics[
    angle=90, width=\columnwidth, trim=2cm 0 1cm 0, clip=true
  ]{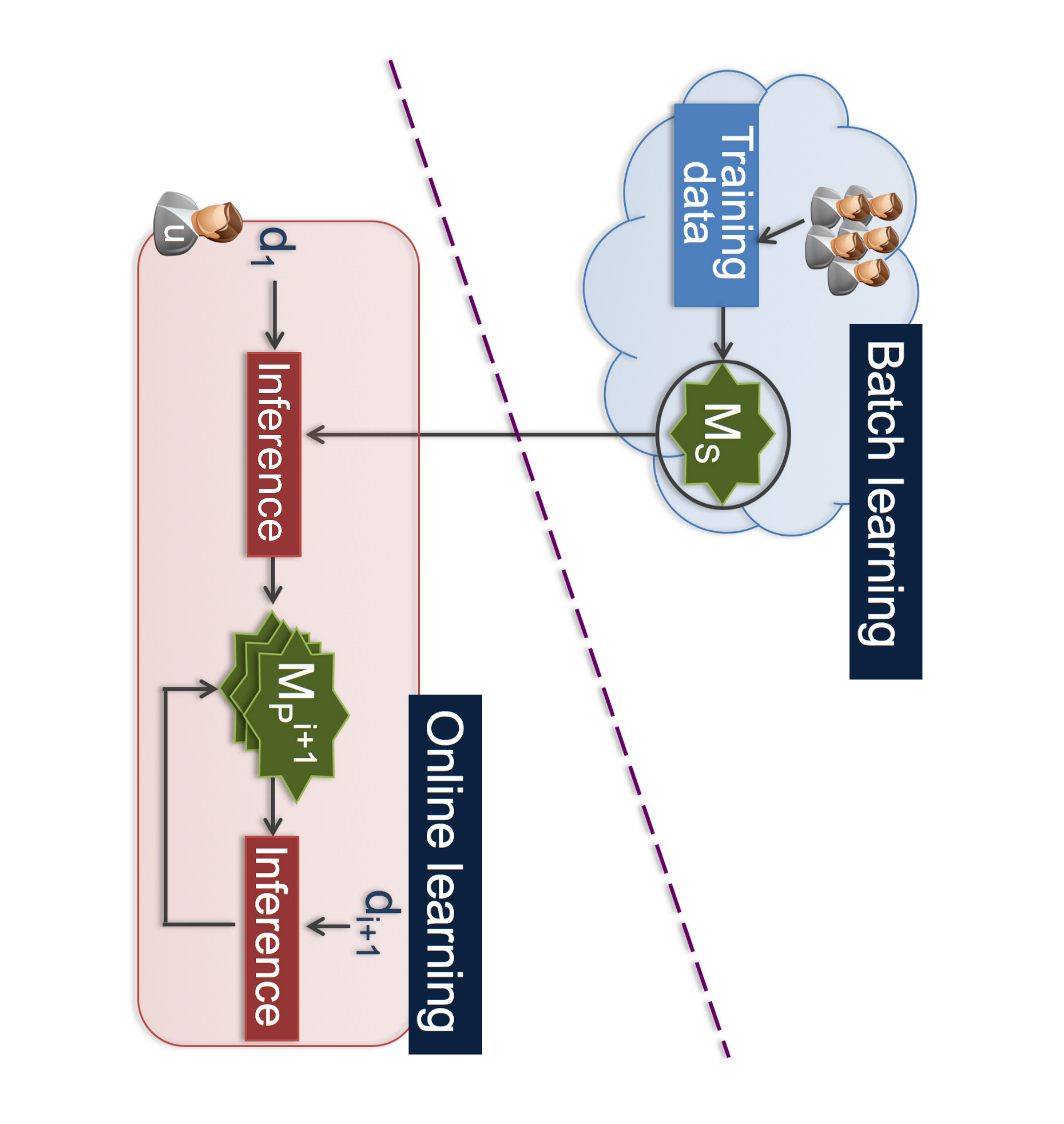}
  \caption{\label{fig:activityRecognitionApp}Our privacy-preserving methodology for activity recognition.}
\end{figure}

We now describe this approach following the overview depicted in Figure~\ref{fig:activityRecognitionApp}. For clarity of exposition, we first describe our approach in the case of supervised learning, taking the activity recognition task we later use in our evaluation as a running example. We then generalise this description to other applications, including our second evaluation example of identifying topics in documents which uses an unsupervised algorithm. This suggests that any learning task, supervised or unsupervised, is amenable to our approach allowing features extracted from users' personal data that they do not wish to disclose to be used to further personalise the initial shared model.

We start by training a \emph{shared} model, $M_S$, to recognise the activity that the user is performing using data sensed with his smartphone's built-in sensors. This \emph{batch learning} is done on a remote server in the cloud using available public data, $d_p$. In the event of not having sufficient public data available for this task, data can be previously gathered from a set of users that have agreed to share their personal data perhaps by providing them with suitable incentives. To assure the confidentiality of their data as well as their presence in the dataset, the \emph{shared} model might be obtained using differentially private training~\cite{shokri2015privacy,mcmahan2016communication,hamm2016learning,papernot2017semi}. 

The user $u$ then obtains the \emph{shared} model from the remote server. With every new sample or group of samples gathered from the smartphone's sensors, the activity that the user is performing is locally inferred using this model. In order to allow for more accurate inferences, the user is prompted to ``validate'' the results by reporting the activity they were performing. The new labelled data so gathered are then used for locally retraining the model, resulting in a new \emph{personal} model, $M_P$.

\subsection{Architecture}
\label{sec:system}

Having described our approach, we now sketch a system architecture that might be used to implement it. This is divided into two parts: \one~residing in the cloud, the first part is responsible for constructing a \emph{shared} model using batch learning; and \two~residing on each individual user's device, the second part tunes the model from the first part using the locally available data, resulting in a \emph{personal} model.

We identify five components in this architecture:

\begin{enumerate}

\item The \textbf{batch training module} resides in the cloud, and is responsible for  training a \emph{shared} model as the starting point using public, or private but shared, datasets that it also maintains. As this component may need to support multiple applications, it will provide a collection of different machine learning algorithms to build various needed models. It may also need to perform more traditional, large scale processing, but can easily be built using modern data processing frameworks designed for datacenters such as Mllib~\cite{meng2016mllib} or GraphLab~\cite{Low:2012:DGF:2212351.2212354}. 

\item The \textbf{distribution module} resides on users' devices and is responsible for obtaining the \emph{shared} model and maintaining it locally. In the case of very large scale deployments, standard content distribution or even peer-to-peer techniques could be used to alleviate load on the cloud service.

\item The \textbf{personalisation module} builds a \emph{personal model} by refining the model parameters of the shared model using the  personal data available on the user's device. This module will also require a repository of different learning algorithms, but the nature of personal computational devices means that there will be greater resource constraints applied to the performance and efficiency of the algorithm implementation.

\item The \textbf{communication module} handles all the communications between peers or those between an individual node and the server. Nodes can register themselves with the server, on top of which we can implement more sophisticated membership management.

\item The \textbf{inference module} provides a service at the client to respond to model queries, using the most refined model available. 

\end{enumerate}

In our implementation, we rely on several existing software libraries to provide the more mundane of these functions, e.g.,~ZeroMQ~\cite{zeromq} satisfies most of the requirements of the communication and model distribution modules, and so we do not discuss these further here.

There are many toolkits, e.g.,~\emph{theano}~\cite{theano} and \emph{scikit-learn}~\cite{scikit}, that provide a rich set of machine learning algorithms for use in the batch training and personalisation modules. However, in the case of the latter, we must balance convenience with performance considerations due to the resource-constrained nature of these devices. In light of this, we use a more recent library, Owl~\cite{owl,liang2017prop}, to generate more compact and efficient native code on a range of platforms, and the source code can be obtained from its Github repository\footnote{https://github.com/ryanrhymes/owl}.

\subsection{Typical Workflow}
\label{sec:workflow}

We briefly summarise the workflow we envisage using  activity recognition as an example.

\begin{enumerate}

\item When the user activates the device for the first time, the device contacts the server and registers itself in order to join the system. The device notices there is no local data for building the model, and sends a request to the server to obtain the \emph{shared} model.

\item After processing the registration, the server receives the download request. The \emph{shared} model has been trained using a initial dataset collected in a suitably ethical and trustworthy way, e.g.,~with informed consent, appropriate compensation, and properly anonymised. The server can either approve the download request, or return a list of peers from whom the requesting user can retrieve the model.

\item Having obtained the \emph{shared} model, the device can start processing  inference requests. At the same time, the device continuously collects user's personal data, in this case, their accelerometer  traces. Once enough local data is collected, the personalisation phase starts, refining the shared model to create a \emph{personal} model.

\item After the \emph{personal} model has been built, the system uses it to serve requests, and continues to refine it as more personal data is collected.

\end{enumerate}

This methodology and hypothesised architecture can be applied to supervised and unsupervised learning tasks in different domains. We next show how it applies to (supervised) activity recognition~(\S\ref{sec:AR}) and (unsupervised) topic modelling~(\S\ref{sec:topicModelling}) using two well-known learning algorithms respectively.

\section{Activity recognition using accelerometer traces}
\label{sec:AR}

In this section, we validate our methodology for supervised learning using a neural network to recognise users' activity using accelerometer traces. Our evaluation for unsupervised learning is detailed in \S\ref{sec:topicModelling}, where we take the task of identifying topics in documents as a case study.

Here we consider a scenario where smartphone users want to train a motion-based activity classifier without revealing their data to others. To test the algorithms, we use the \emph{WISDM} Human Activity Recognition dataset~\cite{kwapisz2011activity}, which is a collection of accelerometer data on an Android phone by $35$ subjects performing 6 activities (\emph{walking}, \emph{jogging}, \emph{walking upstairs}, \emph{walking downstairs}, \emph{sitting} and \emph{standing}). These subjects carried an Android phone in their front pants leg pocket while were asked to perform each one of these activities for specific periods of time. Various time domain variables were extracted from the signal, and we consider the statistical measures obtained for every 10 seconds of accelerometer samples in~\cite{kwapisz2011activity} as the $d = 43$ dimensional features in our models. Our final sample contains $5,418$ accelerometer traces from $35$ users, with on average $150.50$ traces per user and standard deviation of $44.73$.

For the purpose of validation, we compare the performance of the following models:

\begin{itemize}
	\item \emph{Shared}: classifier trained using data from $N-1$ subjects and tested using data from the remaining subject;
	\item \emph{Local}: classifier trained using only data from $1$ subject and tested using also data from the same subject; and
	\item \emph{Personal}: classifier trained using data from $N-1$ subjects (\emph{shared} model), retrained using data from $1$ subject (\emph{local} model) and tested using also data from the latter subject (\emph{personal} model).
\end{itemize}

Thus, we simulated a case where a \emph{shared} model $M_S$ is trained with data from $34$ subjects, while the \emph{personal} model $M_L$ is trained with different samples of data from the remaining participant ($u$). We then compare the accuracy of the \emph{personal} model with the \emph{shared} and \emph{local} models. To this aim, we simulated two other cases: the \emph{shared} model trained using data from $34$ subjects, and the \emph{local} model trained using only local data from $u$. Being $S_u$ the samples of $u$ and $S_r$ the samples of all subjects but $u$, the samples considering for training, validation and test in each model are the following:

\begin{itemize}
	\item \emph{Shared}:
		\begin{itemize}
			\item Training set: $80\%$ of $S_r$
			\item Validation set: $20\%$ of $S_r$
			\item Test set:	$20\%$ of $S_u$
		  \end{itemize}

	\item \emph{Local}:
		\begin{itemize}
			\item Training set: $\{1..60\%\}$ of $S_u$
			\item Validation set: $\{1..20\%\}$ of $S_u$
			\item Test set:	$20\%$ of $S_u$
		  \end{itemize}

	\item \emph{Personal}: We started from the \emph{shared} model and, in order to fine-tune it to obtain the (\emph{personal} model), we considered the same setup as in the \emph{local} model.
\end{itemize}

\subsection{Multi-Layer Perceptron}

We used a Multi-Layer Perceptron as the supervised learning algorithm for recognising activity using accelerometer traces. A Multi-Layer Perceptron or MLP is a type of feedforward Artificial Neural Network that consists of two layers, input and output, and one or more hidden layers between these two layers. The input layer is passive and merely receives the data, while both hidden and output layers actively process the data. The output layer also produces the results. Figure~\ref{fig:mlp} shows a graphical representation of a MLP with a single hidden layer. Each node in a layer is connected to all the nodes in the previous layer. Training this structure is equivalent to finding proper weights and bias for all the connections between consecutive layers such that a desired output is generated for a corresponding input.

The standard back-propagation learning algorithm is used for training the MLP neural architecture. For each accelerometer trace in the training set, weights and bias are modified by computing the discrepancy between the desired and actual outputs and feeding back this error to the inputs, updating the weights and bias in proportion to their responsibility for the output error. The main steps of the back-propagation algorithm are the following:

\begin{enumerate}
	\item \textbf{Initialise the parameters}. All $w_{ij}$'s ($w_{jk}$'s) are initialised to small random values such as the variance of neurons in the network should be $2.0/N$ ($2.0/M$), being $w_{ij}$ ($w_{jk}$) the value of the connection weight between unit $j$ ($k$) and unit $i$ ($j$) in the previous layer, and $N$ ($M$) the number of input (hidden) units~\cite{he2015delving}. The bias, $b_j$ ($b_k$), are initialised to zero.

	\item \textbf{Compute the class scores}. Let the individual components of an input accelerometer trace be denoted by $a_i$, with $i=1,2,...,N$. The output of the neurons at the hidden layer are obtained as: $H_j = \varphi (\sum_{i=1}^N a_i w_{ij} + b_j)$ with $j = 1, 2,..., M$, where $\varphi(\cdot)$ is the activation function and $w_{ij}$ is the weight associated to the connection between the $i$-th input node and the $j$-th hidden node, and $b_j$ the bias. The current recommendation is to use $ReLU$ (Rectified Linear Unit) units~\cite{nair2010rectified, he2015delving} as the activation function ($\varphi(x) = max(0, x)$) though other options are possible. The outputs of the MLP are obtained using $O_k = \varphi ( \sum_{j=1}^M H_j w_{jk} + b_k)$, with $k = 1, 2, ..., C$. Here, $w_{jk}$ is the weight associated to the connection between the $j$-th hidden node and the $k$-th output node, and $b_k$ the bias.

	\item \textbf{Compute the analytic gradient with back-propagation}. This is done by an iterative gradient descent procedure in the weight space which minimises the total loss between the desired and actual outputs of all nodes in the system. The \emph{delta} terms for every node in the output layer are calculated using $\delta_k^o = (O_k - d_k) \varphi^{\prime} (\cdot)$,  with $k = 1,2,..., C$. Here, $\varphi^{\prime} (\cdot)$ is the first derivative of the activation function. Delta terms for the hidden nodes are obtained by $\delta_j^h = \sum_{k=1}^C (w_{jk} \delta_k^o) \varphi^{\prime} (\cdot)$, with $j = 1,2,..., M$.

	\item \textbf{Performing a parameter update}. That is, adjust the weights and bias according to the \emph{delta} terms and $\eta$, the learning rate parameter. For the weights, this is done using $w_{ij} = w_{ij} - \eta \delta_j^h a_i$ and $w_{jk} = w_{jk} - \eta \delta_k^o H_j$. The bias are updated according to $b_{j} = b_{j} - \eta \delta_j^h$ and $b_{k} = b_{k} - \eta \delta_k^o$.

\end{enumerate}

This process is repeated until the network stabilises (converges).

In order to control the capacity of Neural Networks to prevent overfitting, $\ell_2$-regularisation is perhaps the most common form of regularisation. It can be implemented by, for every weight $w$ in the network, adding the term $\frac{1}{2}\lambda w^2$ to the objective, where $\lambda$ is the regularisation strength. Early-stopping is another mechanism to combat overfitting by monitoring the model's performance on a validation set. A validation set is a set of examples neither used for training nor for testing. During training, if the model's performance ceases to improve sufficiently on the validation set, or even degrades with further optimisation, then the training gives up on much further optimisation.

An important aspect of the MLP is the initialisation of the weights and bias, and here is also the main contribution of our proposal. For the \emph{shared} model, $M_S$, we initialise the weights to random small values and the bias to zero in both layers, as suggested by previous literature~\cite{he2015delving}. However, for training the personal model, $M_P$, we start from the weights and bias of the \emph{shared} model, $M_S$.

\tikzset{%
  every neuron/.style={
    circle,
    draw,
    minimum size=0.8cm
  },
  neuron missing/.style={
    draw=none,
    scale=4,
    text height=0.333cm,
    execute at begin node=\color{black}$\vdots$
  },
}

\begin{figure}[t!]
\begin{tikzpicture}[x=1.3cm, y=1.3cm, >=stealth]

\foreach \m/\l [count=\y] in {1,2,3,missing,4}
  \node [every neuron/.try, neuron \m/.try] (input-\m) at (0,2.5-\y) {};

\foreach \m [count=\y] in {1,missing,2}
  \node [every neuron/.try, neuron \m/.try ] (hidden-\m) at (2,2-\y*1.25) {};

\foreach \m [count=\y] in {1,missing,2}
  \node [every neuron/.try, neuron \m/.try ] (output-\m) at (4,1.5-\y) {};

\foreach \l [count=\i] in {1,2,3,n}
  \draw [<-] (input-\i) -- ++(-1,0)
    node [above, midway] {$I_\l$};

\foreach \l [count=\i] in {1,m}
  \node [above] at (hidden-\i.north) {$H_\l$};

\foreach \l [count=\i] in {1,c}
  \draw [->] (output-\i) -- ++(1,0)
    node [above, midway] {$O_\l$};

\foreach \i in {1,...,4}
  \foreach \j in {1,...,2}
    \draw [->] (input-\i) -- (hidden-\j);

\foreach \i in {1,...,2}
  \foreach \j in {1,...,2}
    \draw [->] (hidden-\i) -- (output-\j);

\foreach \l [count=\x from 0] in {input, hidden, output}
  \node [align=center, above] at (\x*2,2) {\l \\ layer};

\end{tikzpicture}
\caption{The architecture of the two-layer feed-forward network.}
\label{fig:mlp}
\end{figure}
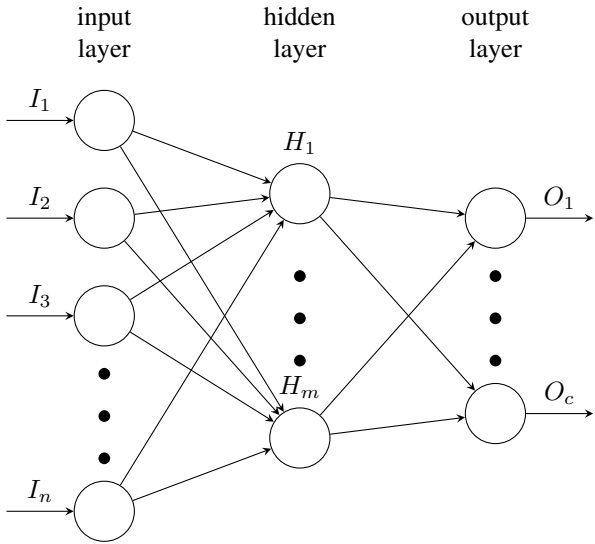

\subsection{Experimental setup}
\label{sec:ARsetup}

We set up a Multilayer Perceptron with $2$ layers for activity recognition, including $1$ hidden layer with $128$ nodes and $1$ logistic regression layer, resulting in $6,406$ parameters to be determined during training. We construct the input layer using the statistical measures of users' accelerometer traces. Because of the sensitivity learning stages to feature scaling~\cite{hinton2010practical} we normalise all statistical measures to have zero mean and unit standard deviation. In the output layer each unit corresponds to an activity inference class, such that unit states can be interpreted as posterior probabilities.

All training procedures were implemented in python using the \emph{Theano} deep learning library~\cite{theano}. The training and testing were performed with $5$-fold cross validation, using early stopping as well as $\ell_2$-regularisation to prevent overfitting. Each neuron's weight in the \emph{shared} and \emph{local} models was initialised randomly from $\mathcal{N}(0,1) / \sqrt{2.0/n}$, where $n$ is the number of its inputs, and biases were all initialised to zero. Parameters in the \emph{personal} model were initialised to the values obtained in the \emph{shared} model. Finally, we used grid search to determine the optimal values of the hyper-parameters, setting the learning rate to $0.05$ for the \emph{shared} model and to $0.001$ for the \emph{local} and \emph{personal} models, and the $\ell_2$-regularisation strength to $1e^{-5}$ for all the models. The training epochs were set to $1000$ in all models, while the batch size was set equal to the size of the training sets in the \emph{shared} model, and to $1$ (online learning) in the \emph{local} and \emph{personal} ones. The reasons behind this are the small size of the dataset, and the availability of the training samples in a real scenario (samples for the \emph{shared} model can be assumed to be all available for training, whereas samples in the \emph{local} and \emph{personal} models become available for training as time goes by).


We repeated the experiment for each participant, using 5-fold cross-validation and different number of samples to train the \emph{local} and \emph{personal} models. In each simulation of every user, we incremented in $1$ the number of samples used for training, and also incremented in $1$ the samples used for validation until reaching $60\%$ of samples for training and $20\%$ for validation, respectively.

\subsection{Results}
\label{sec:ARresults}

\begin{figure}[t!]
\centering
\includegraphics[angle=0, scale=0.4]{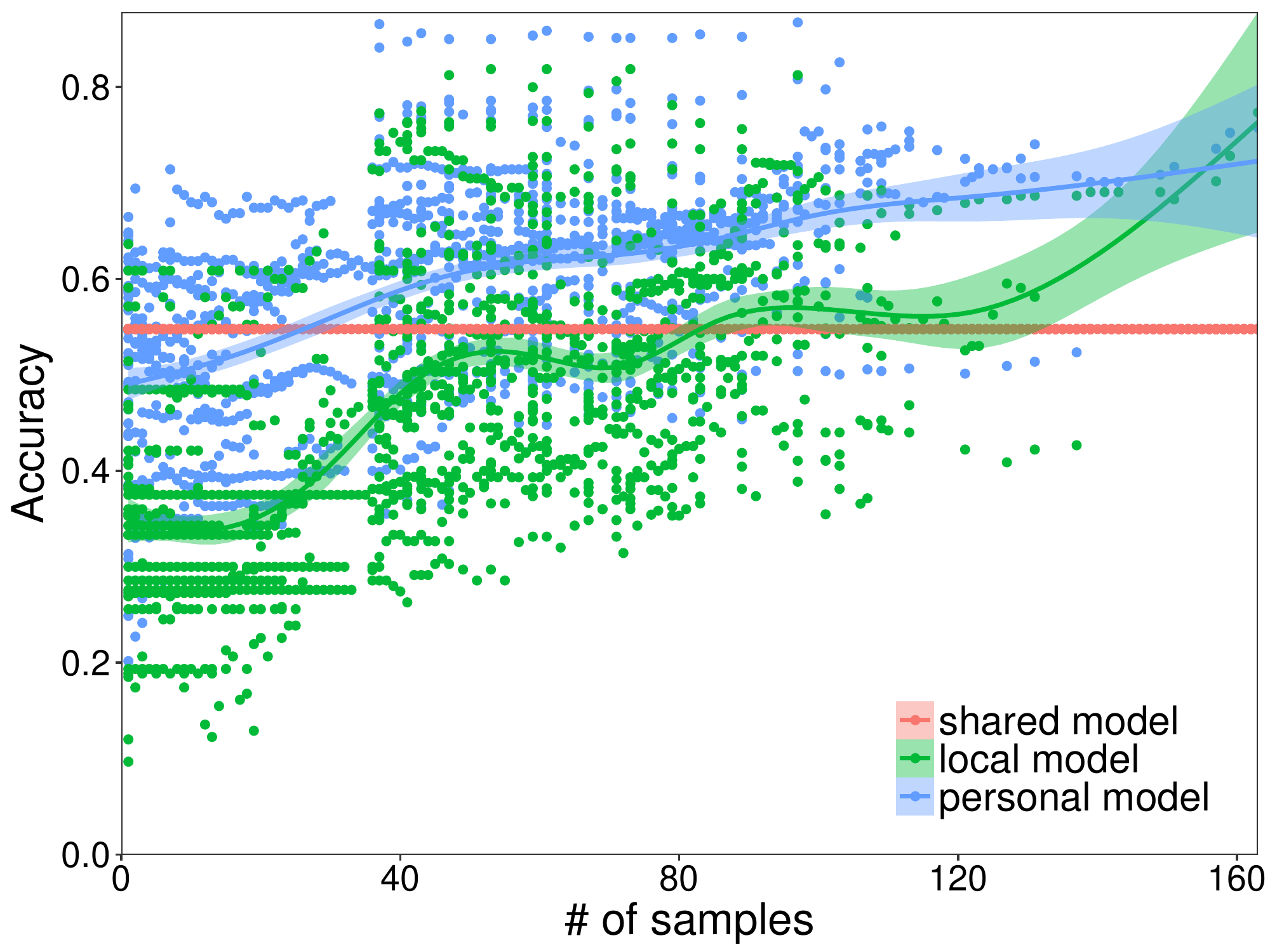}
\caption{Accuracy obtained with each model for different number of local samples per user.}
\label{fig:activityRecognitionResults}
\end{figure}

Figure~\ref{fig:activityRecognitionResults} reports the accuracy achieved with each model when considering different number of local samples per user. Results show that the effect of training or retraining a model with few samples from the individual under test produces worse predictions than using samples from other individuals (\emph{shared} model). That is, while the model is adapting to the new \emph{scenario}, the performance of the prediction slightly drops. However, when more samples ($20$ on average or more) are used to retrain this \emph{shared} model, the accuracy of the prediction exceeds the accuracy obtained with the \emph{shared} model itself. Specifically, the accuracy increases with increments in the number of samples used for retraining the model. That is, the more local samples considered to retrain the model, the more \emph{personalised} it becomes for the considered individual. However, although the improvement on the accuracy with the increment of the number of samples is also shared with the \emph{local} model, more samples per individual are required for training a model from scratch (\emph{local} model) in order to obtain the same accuracy than when starting from a \emph{shared} model (\emph{personal} model). We also observe that, after on average $163$ samples, the \emph{local} model performs better than the \emph{personal} model. This means that the user would need to perform and label, on average, $163$ activities in order to get a \emph{local} model that outperforms her \emph{personal} one. However, this is not significant, since there is one unique user in the dataset with that number of samples or higher available for training. In summary, \one~retraining a \emph{shared} model locally using $20$ or more samples from the user increases the accuracy with respect to that obtained with the \emph{shared} model, and \two~to obtain the same accuracy when training a model from scratch using only local samples, more than $150$ training samples are required on average.

\section{Topic modelling of personal text corpus}
\label{sec:topicModelling}

Text corpus, such as web pages, documents, e-books and emails, is one of the most widely distributed media on the Internet and also dominates many users' devices. Text classification therefore has a wide application to facilitate people's daily life by grouping or filtering the items in a text corpus based on certain topics. Because personal text corpus often contains a significant amount of private information, uploading such corpus to a public cloud will certainly breach user privacy.

Consider now a scenario where users wish to classify the textual documents on their computers without revealing their content to others. Being more precise, we consider researchers working on a confidential project within a company that do not wish to disclose the publications they are reading, but they do wish to have the documents classified according to their content. To simulate such a scenario, we use two text datasets in our evaluation: the NIPS~\cite{nips} and the Wikipedia~\cite{wikipedia} datasets. The NIPS dataset is a collection of papers published in NIPS conference over the past two decades. It contains about $1.5$k papers and $1.9$~million words. For Wikipedia, we download its latest English dump in January 2017 which contains about $5$ million articles and $2.9$~billion words.

For the purpose of validation, we compare the performance of the following models:

\begin{itemize}
	\item \emph{Local}: topic extraction using only data from the NIPS dataset (local data);
	\item \emph{Personal}: topic extraction using only data from the Wikipedia dataset (\emph{shared} model), and retraining using data from the NIPS dataset (\emph{local} model).
\end{itemize}

Thus, we simulated a case where a \emph{shared} model $M_S$ is trained with data from the Wikipedia dataset, while the \emph{personal} model $M_L$ is trained with the NIPS dataset. We then compare the accuracy of the \emph{personal} model with the \emph{local} model. To this aim, we also simulated the case where the \emph{local} model trained using only local data from the NIPS dataset.

\subsection{Latent Dirichlet Allocation}
\label{sec:topicModelling:lda}

A \textbf{topic model} is an effective tool for text mining. It aims to construct a statistical model to represent the abstract ``topics'' contained in a collection of documents. By doing so, similar documents can be grouped together for future queries. As each document is composed of a sequence of words, it is often represented as a very high-dimensional and sparse vector using a bag-of-words model. Each dimension represents one unique vocabulary in the dictionary extracted from the corpus, and the magnitude of each dimension is often calculated as the term frequency in the corresponding document. Therefore, the dimensionality of these vectors depends on the size of dictionary, and it is common the dimensionality is over dozens of thousands.

\textbf{LDA} is a generative model which explicitly models topics as latent variables based on the co-occurrences of terms and documents in a text corpus. LDA is similar to pLSA~\cite{Hofmann:1999:PLS:312624.312649} but replaces the maximum likelihood estimator with Bayesian estimator, hence it is sometimes referred to as the Bayesian version of pLSA. LDA assumes that the topic distribution has a Dirichlet prior.

As each document can be represented as a vector, finding a given document's similar documents is equivalent to search for its $k$-nearest neighbours in the high-dimensional space. Many prior works~\cite{7462177, Hajebi:2011:FAN:2283516.2283615, He:2010:SSS:1835804.1835946, 7840682} focus on building compact and efficient data structure and search algorithm to speed up the queries to the models.

As mentioned, LDA is an unsupervised learning method and its goal in training is to maximise its likelihood function. Its model contains two important parameters:

\begin{itemize}

\item\textbf{Document-Topic distribution}: it indicates the probability distribution of each document over a set of topics.

\item \textbf{Topic-Word distribution}: it indicates the probability distribution of each topic over a set of words extracted from the text corpus.

\end{itemize}

Essentially, the two parameters are represented as two matrices in the algorithm containing the information of document-topic and topic-word assignment respectively. A typical training task can be divided into two phases: First, the two parameters will be initialised by assuming both have a Dirichlet prior; second, the model will be updated by applying Collapsed Gibbs Sampling to all the documents. The second phase will be applied iteratively until we reach the pre-defined number of iterations or the model converges. To evaluate the effectiveness of an LDA model, we can measure the log likelihood of the constructed model over a test data set. The log likelihood indicates how well the model can interpret the given data set, and the higher value it is, the better it is.

\subsection{Experiment setup}
\label{sec:topicModelling:setup}

In our topic modelling scenario, a user $u$ owns a set of text documents, $D_u$, and wants to identify their topics without revealing their content. There is another set of publicly available documents, $D_r$, that he can benefit from. The set of $u$'s documents, $D_u$, is composed by the documents in the NIPS dataset, whereas $D_r$ is formed by different random samples of documents from the Wikipedia dataset. 

We start by building a \emph{shared} LDA model called $M_S$ out of the public available documents $D_r$ as we did in the previous activity recognition case. However, one thing worth noting here is that $M_S$ only includes the Topic-Word distribution (as well as a dictionary to tokenise the documents) which is a very sparse matrix. The Document-Topic parameter depends on the specific text corpus and is not useful for others to initialise a \emph{personal} model, hence it is not necessary to include into $M_S$.

We then build the new LDA model, \emph{personal} model or $M_P$, using the new document samples that corresponds to users' personal data, i.e., $D_u$, and compare it with the \emph{local} model trained solely using user's documents. Specifically, the method for building the \emph{local} model is just repeating the typical process of building the model from scratch. Alternative method, i.e. the method to build the \emph{personal} model, is to request $M_S$ and use it to initialise the local Document-Topic distribution parameter instead of assuming a Dirichlet prior. In the following, we will compare and present the accuracy and efficiency of the \emph{local} and \emph{personal} models.

We use the topic modelling module in Owl~\cite{owl} library to perform the aforementioned experiments. Each experiment is repeated 10 times to guarantee its consistency.

\subsection{Improved accuracy and efficiency}
\label{sec:topicModelling:result}

\begin{figure}[t!]
\centering
\includegraphics[scale=0.65]{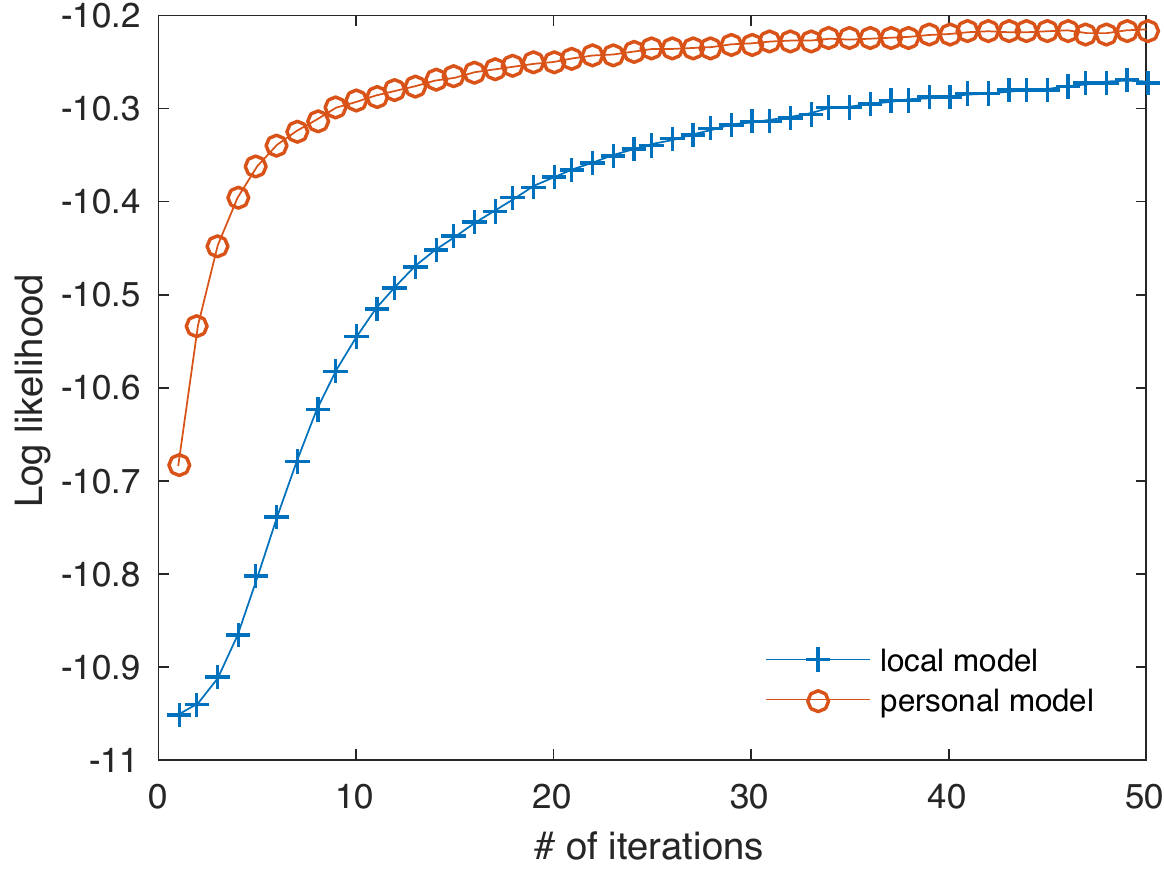}
\caption{\label{fig:lda:01}Log likelihood in each iteration while building a LDA model with and without using a \emph{shared} model $M_S$. That is, for building the \emph{personal} and \emph{local} LDA models.}
\end{figure}

Figure~\ref{fig:lda:01} presents the evolution of log likelihood in each iteration while building the \emph{personal} ($M_P$) and \emph{local} models. NIPS dataset is used in this first experiment, and the \emph{shared} model, $M_S$, is trained using 50\% of the data. A user's local data are generated by randomly selecting 300 documents from the rest of the dataset.

The two lines in the figure correspond to the two ways of building the user's topic model: the red line is for using $M_S$ to initialise model (\emph{personal} model) whereas the blue line is for building the model from scratch by solely using the user's local documents (\emph{local} model). As we see, the \emph{personal} model is able to achieve higher likelihood in each iteration than the \emph{local} one. Meanwhile, it also indicates that the model is able to converge much faster given a target accuracy.

\begin{figure}[t!]
\centering
\includegraphics[scale=0.65]{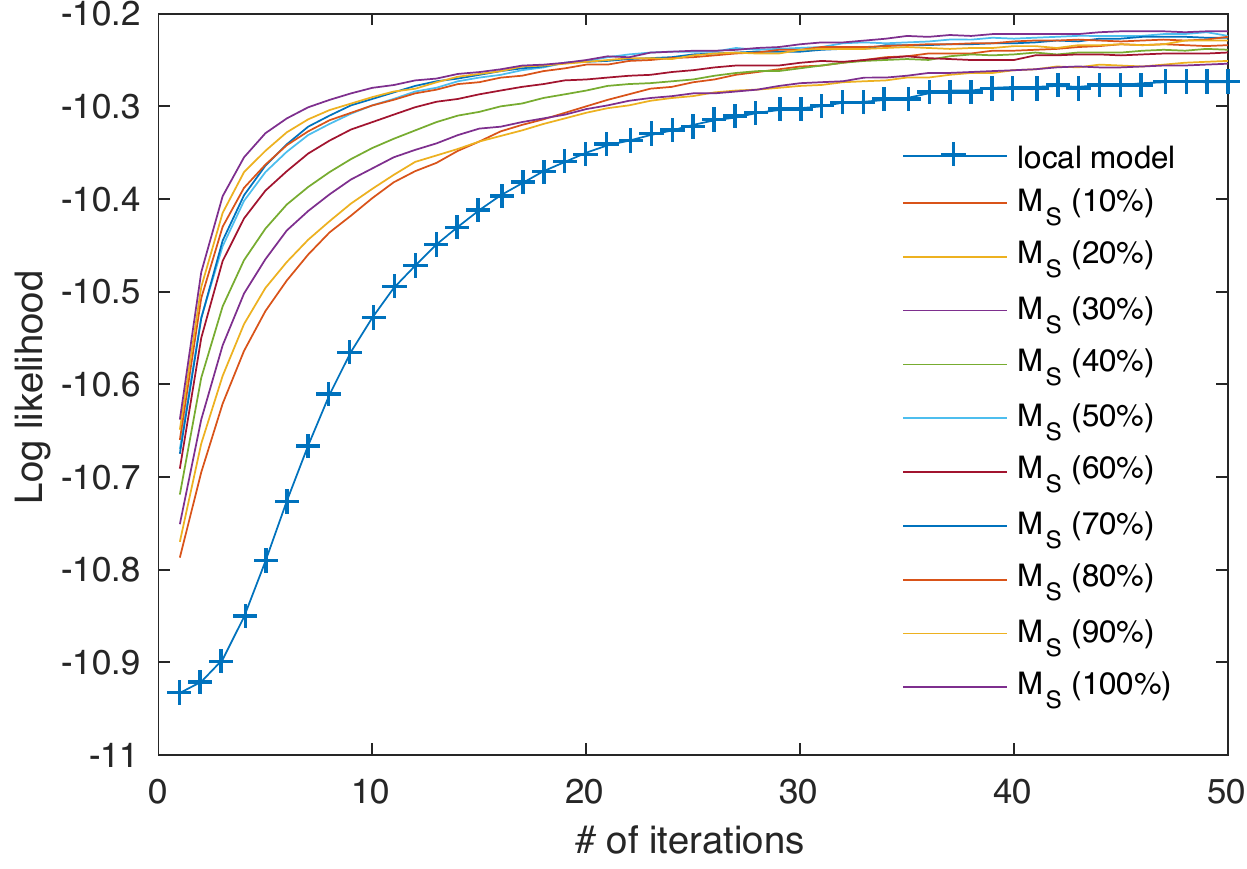}
\caption{\label{fig:lda:02}Log likelihood in each iteration while building a \emph{personal} LDA model by including different amount of public data into the shared model $M_S$.}
\end{figure}

The NIPS dataset is larger than the activity recognition dataset, which allows us to raise and answer the next question: \emph{how much public data do we need to use for building the \emph{shared} model $M_S$?} This question becomes very relevant especially when the amount of public or shared data is limited. In the next experiment, while keeping the rest of the experiment settings the same, we build multiple \emph{shared} models $M_S$ by increasing the amount of documents for training step by step, from 100 to 1000. With these new $M_S$, we repeat the same experiment presented in Figure~\ref{fig:lda:01} to investigate how the amount of public data used to train $M_S$ impacts the personal training (\emph{personal} model). Figure~\ref{fig:lda:02} presents our results. The blue line with ``$+$'' marker at the bottom is the same as that in Figure~\ref{fig:lda:01}, representing the training without using $M_S$ (\emph{local} model). The rest of the lines represent the log likelihood of the $M_S$ using different percentage of the 1000 documents as the public sample. All $M_S$ models are trained using a fixed number (i.e.,~50) of iterations.

Results in Figure~\ref{fig:lda:02} show that including more data in $M_S$ will certainly improve the efficiency and accuracy when localising or personalising the \emph{shared} model which is reflected as gradually improved likelihood in each line. Note that including more data in training $M_S$ will not increase the parameter size therefore it does not introduce extra overhead in distributing the \emph{shared} model. However, these results also deliver another important message, namely such benefits drop quickly as we add more and more data. The most significant improvement appears in the very beginning when we only include a small amount of data (i.e., 10\%). To some extent, it justifies our proposal by showing a small amount of public data is able to boost local training.

\subsection{Topic-based local dataset}

Wikipedia dataset contains a rich set of meta information such as manually assigned topic categories, which can help us in simulating users who have different interests in various topics. Users of different interest may possess a rather different local dataset from each other. The question is whether our method is still effective when the topics in the local data are only a subset of those included in the documents used for training the \emph{shared} model.

\begin{figure}[t!]
\centering
\includegraphics[scale=0.65]{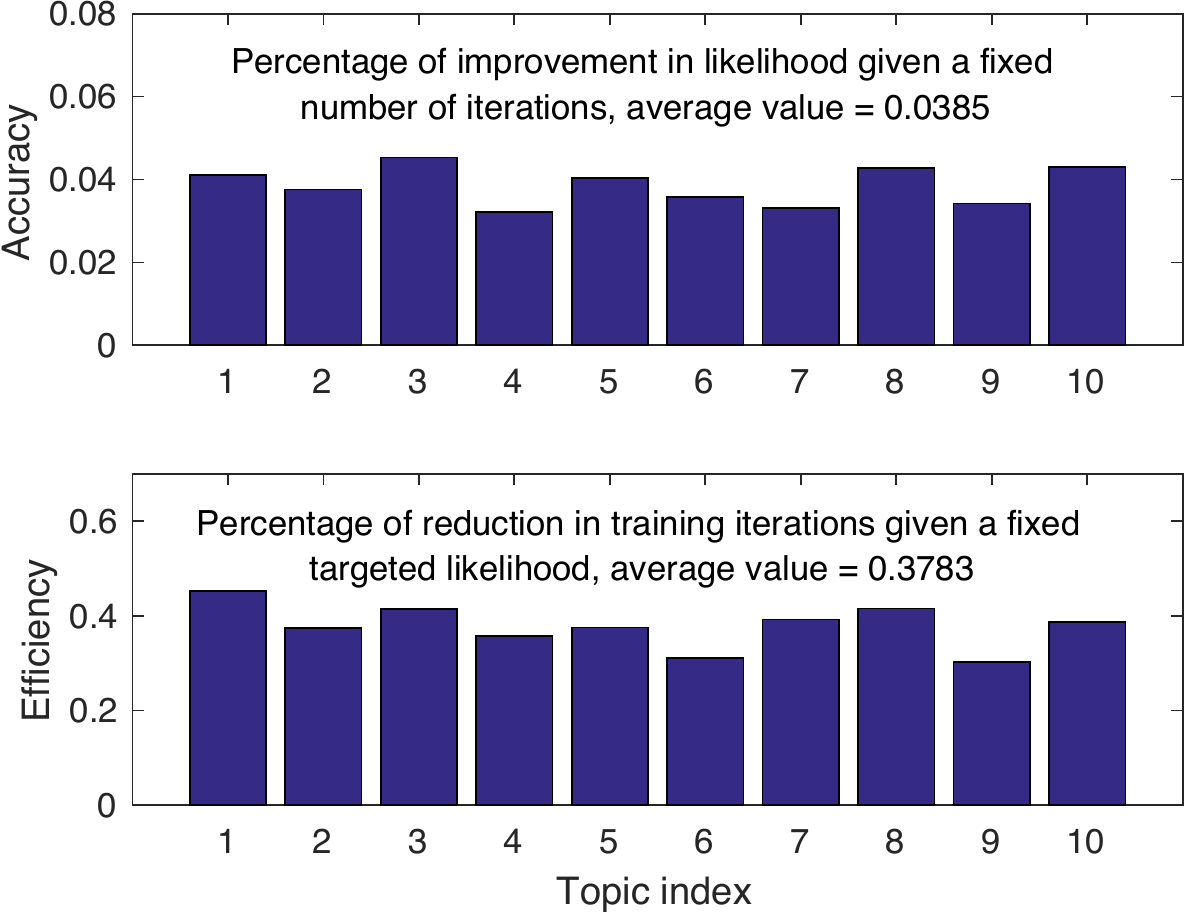}
\caption{\emph{Shared} model also helps in improving accuracy and efficiency even when the local datasets have a strong focus on specific topics. Ten randomly selected topics from Wikipedia are used to generate local dataset. The upper figure presents the percentage of improving log likelihood for a fixed number of iterations, while the lower figure measures the percentage of reduction in training iterations for a targeted likelihood. Both figures have the same x-axis which is indexed by topic numbers.}
\label{fig:lda:05}
\end{figure}

To answer the question, we design other experiment wherein when we generate the local dataset, we only randomly select the articles from a pre-determined topic, e.g.,~computer networking, architecture design, British history. We generate multiple local datasets using different topics. The $x$-axis in Figure~\ref{fig:lda:05} shows the topic indices in the Wikipedia dataset, and we present the results of 10 of those selected topics. The way of selecting the shared data for training $M_S$ remains the same. In total, we randomly sample 1000 articles from the whole Wikipedia as public or shared data, and 500 articles of a given topic for each local dataset.

We first measure the absolute amount of improvement in log likelihood between the first and the tenth iteration while training a \emph{local} model, based on which we then calculate how much we can improve by starting with a \emph{shared} model (\emph{personal} model). The results in upper part of Figure~\ref{fig:lda:05} show that the improvement varies between 3.22\% to 4.53\% with an average equals to 3.85\%. We also measure the improvement in efficiency by investigating how much we can reduce the number of iterations to reach a targeted log likelihood with the help of a \emph{shared} model. In this experiment, we set the targeted log likelihood to $-10.203$ and the lower part of Figure~\ref{fig:lda:05} presents our results. Similar to the upper one, using a \emph{shared} model can significantly boost the training efficiency, with the minimum over 30\% reduction in iterations over all cases. On average, we are able to save over 37.8\% iterations. Another thing worth mentioning here is that since the \emph{shared} model is already sparse, it further reduces the time spent in each iteration than training purely on local data (\emph{local} model) which needs to start with a highly dense local model. This indicates the saving is even more significant in terms of absolute amount of training time reduced.

\section{Practical considerations}
\label{sec:discussion}

In the following, we discuss the privacy guarantees of our methodology and its robustness against adversarial attacks (\S\ref{sec:adversarial}). We then demonstrate the feasibility of its deployment by presenting a performance evaluation on a representative resource-constrained device (\S\ref{sec:deployment}). We conclude this section discussing the applicability and some limitations of our methodology.

\subsection{Adversarial attacks}
\label{sec:adversarial}

Our system can suffer the attacks and consequences of malicious users. There are several potential attacks against any learning system~\cite{barreno2010security, huang2011adversarial}. Here we focus on how privacy and causative attacks might affect our system. On a privacy attack the adversary obtains information from the learner, compromising the secrecy or privacy of the system's users. The aim of a causative attack is on altering the parameters of the target model by manipulating the training dataset. An example of this type of attacks are poisoning attacks, where an attacker may poison the training data by injecting carefully designed samples to eventually compromise the whole learning process. The target model then updates itself with the poisoned data and gradually compromises. Below we describe the potential effects of these attacks in our system.

\subsubsection{Privacy attacks}
\label{sec:privacy}

Our solution guarantees the confidentiality of users' data (potential users) given that their devices are not compromised, since their personal data never leave their devices. Since both the data and the personal model reside on the user's device, attacks such as model inversion~\cite{fredrikson2014privacy} --where an attacker, given the model and some auxiliary information about the user, can determine some user's raw data; and membership query~\cite{shokri2017membership}, where, given a data record and black-box access to a model, an adversary could determine if the record was in the model’s training dataset, cannot affect our users. However, we cannot assure the confidentiality of the data, neither robustness against these attacks, for those users that have freely aggreed to share their data in the same way as the big corporations are not doing so with their customers data. For many applications we envisage and describe in the introduction, such as those based on object recognition or those that work with textual data, there is already a large amount of data freely available on the Internet with which to build the shared model, and whose confidentiality does not need to be guraranteed. On the other hand, for applications such as face or speaker recognition, techniques based on differentially private training~\cite{shokri2015privacy,mcmahan2016communication,hamm2016learning,papernot2017semi} could be applied in order to, a priori, guarantee the confidentiality of the volunteers' data. On the contrary, the training of the personal model for the final users happens locally on their devices so that neither their data nor their personal model leave their devices, and its confidentiality is guaranteed by the security offered by their device, security that is out of the scope of the methodology proposed here.

\subsubsection{Poisoning attacks}
\label{sec:poisoning}

We envisage two different points or steps in our system that adversaries might wish to attack: when building the shared model in a remote server in the public cloud using public data available or \emph{shared} by a group of volunteers, and when personalising the model by local retraining in the user's device (\emph{personalisation}). In the case of a poisoning attack to our proposed methodology, the shared model can be corrupted by malicious volunteers poisoning the data with fake samples. However, during the local retraining, if the adversary wishes to corrupt the personal model, he needs to gain access to the local device of the user to poison the data and fool the model. Poisoning the data to train the personal model needs the attacker to gain access to the local device of the user.
\subsection{Deployment Feasibility}
\label{sec:deployment}

The success of our privacy-preserving methodology for learning personal models is conditional on the ability to run model refinement in \textit{near} real-time on resource-constrained personal devices which lack the capabilities of cloud-based servers. These environments are of increasing interest for deployment of such techniques as availability of computation resources outside datacenters continues to increase with creation of ``fog computing'' environments using cheap and energy-efficient platforms such as those based on ARM processors~\cite{fog}.

To verify the deployment feasibility of our approach on resource-constrained personal devices we evaluate, for both tasks, the second phase of our approach, testing the refinement and inference aspects on a Raspberry Pi 3 Model B~\cite{raspberrypi} as representative of these sorts of environments.

\begin{figure}[t!]
  \centering
  \includegraphics[angle=0, scale=0.4]{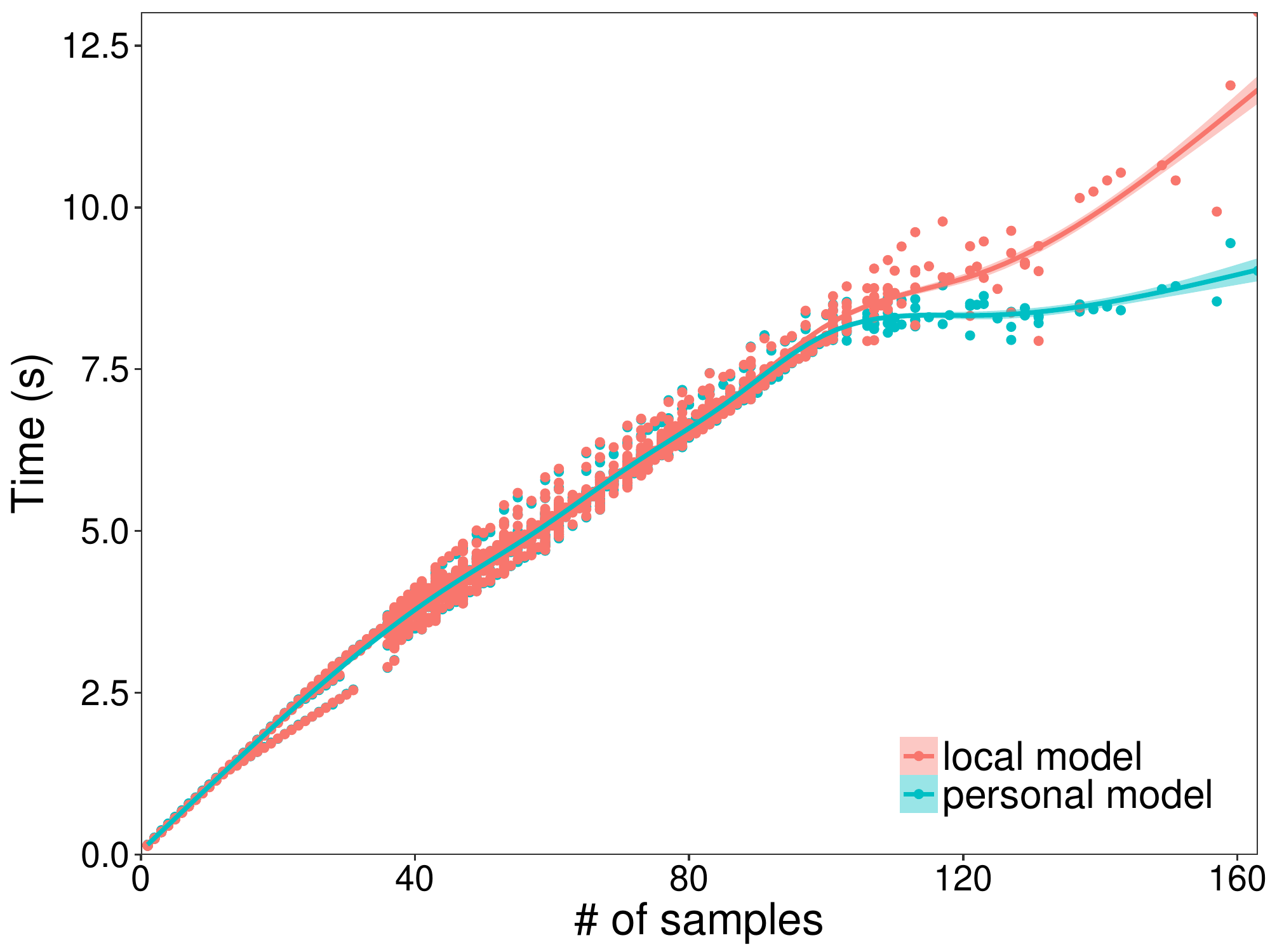}
  \caption{\label{fig:ActivityRecognitionTime}Time for training with each model, varying the number of local samples per user.}
\end{figure}

Figure~\ref{fig:ActivityRecognitionTime} shows the time taken for training a \emph{personal} model (refining the initial \emph{shared} model), and for the alternative approach of learning a \emph{local} model using only a single user's data available locally, for the setup in \S\ref{sec:AR}. In both cases training takes seconds to complete, with time increasing linearly with the number of samples considered due to the online nature of the training process (only one sample is considered at every update of the model). The time for making the inference is insignificant compared with the time for training, being on the order of milliseconds.

\begin{figure*}[t!]
  \centering
  \includegraphics[scale=0.57]{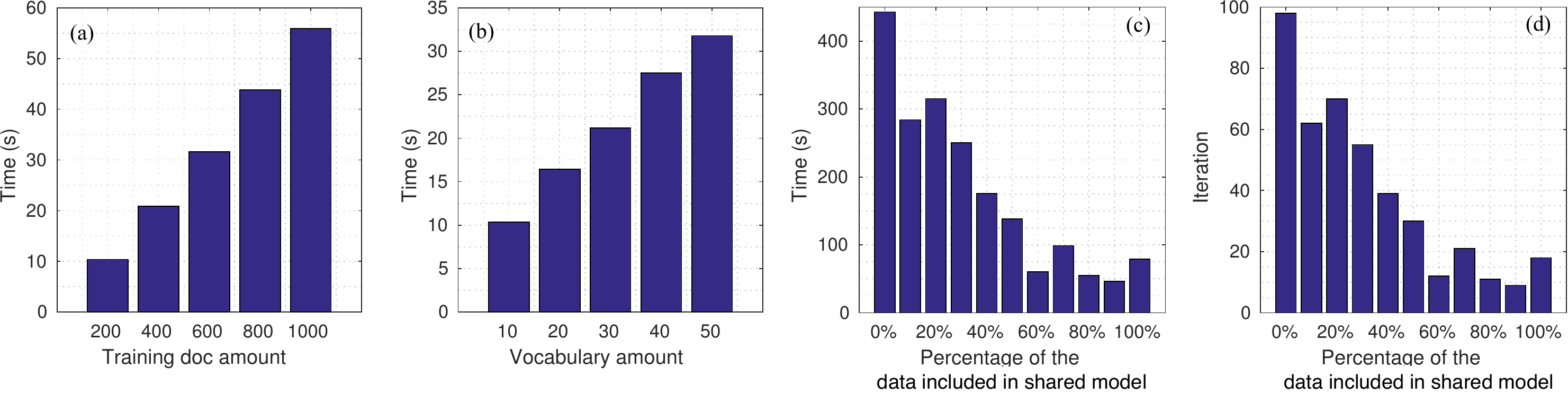}
  \caption{\label{fig:lda:03} (a-b) Time spent per iteration on building the model with two variables: the number of training documents and the vocabulary size; (c-d) time taken to reach a given likelihood (-10.230) started from different \emph{shared} models, with different percentage of data included in the initial \emph{shared} model.}
\end{figure*}

Figure~\ref{fig:lda:03} reproduces results from \S\ref{sec:topicModelling} on the same resource-constrained device using the NIPS dataset. Figures~\ref{fig:lda:03}(a-b) show the time spent per iteration while building the model, respectively varying the number of training documents while keeping the vocabulary size fixed, and the vocabulary size while keeping the number of training documents fixed. Each test is repeated for 20 iterations. In both cases, we observe the expected linear growth in model training time.

Figures~\ref{fig:lda:03}(c-d) show the time it takes to reach a certain level of likelihood (we use -10.230 as the threshold, as in the previous experiments) given different \emph{shared} models. All models are trained with 200 documents. The results align with what we have found out in Figure \ref{fig:lda:02}: including more data decreases both the time and number of iterations required to reach the desired precision threshold. This decrease is most obvious when the model moves from not using any data to including a small amount (10\%). As further data is included, the benefits quickly diminish.

\subsection{Applicability and limitations}
\label{sec:applicability}

Our privacy-preserving methodology for learning analytics may be used for different applications beyond the activity recognition and topic modelling tasks described and evaluated in~\S\ref{sec:AR} and~\S\ref{sec:topicModelling}, respectively. This includes applications that use more complex data than text and numbers, such as object, face and speaker detection from multimedia data. Beyond the limitations of latency and energy consumption of the final devices to cope with very large models, the complexity of the data and the degree of dependency between the data and the task play a key role in the applicability of such methodology. For example, the metholodoly is expected to be effective in detection (object, face, speaker, etc.), but not for recognition where the \emph{local} classes to infer are different from the \emph{shared} ones (different faces, different speakers, etc.).

Other limitation of our methodology is that it seeks to assure the confidentiality of end users' data, but its privacy guarantees do not cover those subjects who volunteered their data to train the \emph{shared} model. Potential extensions of our methodology might include the use of mechanisms such as differentially private training to obtain the \emph{shared} model, guaranteeing the confidentiality of volunteers' data as well as their presence in the dataset.

\section{Related Work}
\label{sec:related}

Data-driven solutions are now pervasive in areas such as advertising, smart cities and eHealth~\cite{PhilipChen2014314, 6547630}. Almost everything we do in our daily lives is tracked by some means or another. Although the careful analysis of these data can be highly beneficial for us as individuals and for the society in general, this approach usually entails invasion of privacy, a high price that progressively more people are not willing to pay~\cite{brandimarte12013misplaced}.

Several privacy-preserving analytical solutions have been proposed to guarantee the confidentiality of personal data while extracting useful information~\cite{agrawal2000privacy,aggarwal2008general,erkin2013privacy,bellala2016securing}. Prominent among them are those that build on Dwork's \emph{differential privacy} framework~\cite{sarwate2013signal,song2013stochastic,chaudhuri2011differentially,fredrikson2014privacy,abadi2016deep}, which formalises the idea that a \emph{query} over a sensitive database should not reveal whether any one person is included in the dataset~\cite{dwork2008differential}. In the case of machine learning, the idea is that a differentially-private model should not reveal whether data from any one person were used to train the model. Most of these techniques for differentially-private machine learning are usually based on adding noise during the training, which leads to a challenging trade-off between accuracy and privacy. 

Distributed machine learning (DML) arised as a solution to well utilize large computer clusters and highly parallel computational architectures to speed up the training of big models over the large amounts of data available nowadays~\cite{li2014communication}. Systems dealing with very large datasets have already had to handle the case where no single node can contain and process the entire dataset, but the dataset and/or the model to learn are paralellised among different machines, models are sequentially trained on each single machine, and some sort of synchronisation mechanism is applied to aggregate the parameters of the model to learn~\cite{cormen1996bridging,agarwal2011distributed,ho2013more,low2012distributed}. DML may also be a potential solution when the volume of the data is not the main issue, but the distribution occurs when different entities own different datasets which, if aggregated, would provide useful knowledge. However, the sensitivity of such data often prevents these entities from sharing their datasets, restricting access to only a small set of selected people as in the case of patients' medical records~\cite{bellala2016securing}. Several solutions have been proposed for privacy-preserving distributed learning, where information is learnt from data owned by different entities without disclosing either the data or the entities in the data. Shokri and Shmatikov~\cite{shokri2015privacy} and McMahan \etal~\cite{mcmahan2016communication} propose solutions where multiple parties jointly learn a neural-network model for a given objective by sharing their learning parameters, but without sharing their input datasets. A different approach is proposed by Hamm~\etal~\cite{hamm2016learning} and Papernot \etal~\cite{papernot2017semi}, where privacy-preserving models are learned locally from disjoint datasets, and then combined on a privacy-preserving fashion. However, the privacy guarantees of some of these solutions have recently been called into question~\cite{hitaj2017deep}.


Contrary to previous approaches, our aim is not on learning a global model from sensitive data from multiple parties, but to learn a personalised model for each individual party that builds on a model learnt from a relatively small set of others parties, without requiring access to their raw data. We build \emph{personal} learning models similar to the personal recommender system by Balasubramanian \etal~\cite{balasubramanian2017pria} but generalising the solution to any learning algorithm. Our solution takes advantage of transfer learning~\cite{pan2010survey} to achieve better performance than algorithms trained using only local data, particularly in those common situations where local data is a scarce resource. This solution brings most of the data processing to where the data resides and not the other way around, exactly as the edge computing paradigm calls for~\cite{shi2016edge}. Recent work have demonstrated the feasibility of running complex deep learning inferences on local devices such as smarphones~\cite{georgiev2014dsp,georgiev2016leo}. While in these works models are previously trained in an offline manner, our experiments in \S\ref{sec:deployment} proved that both the inference and the local retraining can be performed locally on a low-power device such as the Rasperry Pi in a timely manner.

\section{Conclusion}
\label{sec:conclusion}

Our privacy-preserving methodology for learning analytics relies on the ability of current personal devices such as smartphones, tablets and small form-factor computers such as the Raspberry Pi to carry out traditionally resource-demanding tasks. By splitting model training between the cloud and the personal device, we avoid sending personal data to untrustworthy remote entities in the cloud while maintaining efficiency and improving accuracy of both training and inference. Users thus keep all rights over their personal data while retaining the benefits of learning-based services. 

We demonstrated our methodology for two typical learning tasks, one supervised and one unsupervised: activity recognition from accelerometer data, and identification of topics in text documents. Our experiments showed improvements both in accuracy and efficiency using this methodology with respect to both traditional cloud-based solutions and solutions based on training the model using \emph{only} the data available from the user to whom we are providing the service. We also demonstrated the feasibility of our approach by examining performance of implementation of the local step of our methodology on a Raspberry Pi 3 Model B.

Our results prove that this approach is promising, and we believe that it is widely applicable and able to positively improve the privacy-preservation of applications where data is distributed between entities whose confidentiality needs to be secured. In a world where the Internet of Things is connecting progressively more devices everyday -- some of which already reside in our home, and the rising concerns about the possibility of private data leaving or getting stolen from our smart homes, our privacy-preserving methodology comes to the aid of designers and developers of smart home applications to fulfil the need of privacy guarantees that their users demand. However, there are certainly areas appropriate for future work. For example, our current solution does not leverage any cooperation between user nodes, and we plan to explore a \emph{cooperative learning} approach to allow \emph{similar} users to jointly train machine learning models in cases where they do not individually have sufficient labelled data locally to retrain a \emph{shared} model, likely the case during the first moments they use the learning service. A potential solution to boost the accuracy may deal with similar users retraining the \emph{shared} model collaboratively, in a privacy-preserving fashion.

\bibliographystyle{IEEEtran}
\bibliography{iotdi18}

\end{document}